\documentclass{article}

\usepackage[preprint]{neurips_2025}

\usepackage{graphicx}
\usepackage{booktabs}
\usepackage{longtable}
\usepackage{multirow}
\usepackage{caption}
\usepackage[ruled,vlined]{algorithm2e}
\usepackage{fancyvrb}
\usepackage{tcolorbox}
\usepackage{array}
\usepackage{paralist}
\usepackage{amsmath, amssymb}
\usepackage{microtype}
\usepackage{hyperref}

\title{ConTextual: Improving Clinical Text Summarization in LLMs with Context-Preserving Token Filtering and Knowledge Graphs}

\author{%
  Fahmida Liza Piya \quad Rahmatollah Beheshti \\
  University of Delaware\\
 \texttt{lizapiya@udel.edu} \quad \texttt{rbi@udel.edu} \\
}

\begin{document}
\maketitle

\begin{abstract}
Unstructured clinical data can serve as a unique and rich source of information that can meaningfully inform clinical practice.
Extracting the most pertinent context from such data is critical for exploiting its true potential toward optimal and timely decision-making in patient care. While prior research has explored various methods for clinical text summarization, most prior studies either process all input tokens uniformly or rely on heuristic-based filters, which can overlook nuanced clinical cues and fail to prioritize information critical for decision-making. 
In this study, we propose \texttt{Contextual}, a novel framework that integrates a Context-Preserving Token Filtering method with a Domain-Specific Knowledge Graph (KG) for contextual augmentation. By preserving context-specific important tokens and enriching them with structured knowledge, \texttt{ConTextual} improves both linguistic coherence and clinical fidelity. Our extensive empirical evaluations on two public benchmark datasets demonstrate that \texttt{ConTextual} consistently outperforms other baselines. 
Our proposed approach highlights the complementary role of token-level filtering and structured retrieval in enhancing both linguistic and clinical integrity, as well as offering a scalable solution for improving precision in clinical text generation\footnote{Our code repository is publicly available at: \url{https://github.com/healthylaife/ConTextual.git}.}.

\end{abstract}

\section{Introduction}
Electronic health records (EHRs) are central to modern medical informatics, providing a rich repository of structured and unstructured data that drives clinical decision-making and research~\citep{piya2024healthgat, wornow2023ehrshot,poulain2022few}. While structured data enables systematic analyses, unstructured components—such as discharge summaries and progress notes—contain nuanced clinical insights that are difficult to summarize due to their reliance on complex medical terminology, subtle contextual cues, and intricate interrelationships~\citep{hossain2023natural}. Efficiently extracting and summarizing these insights is critical for improving patient care~\citep{piya2025advancing}, yet it remains an open challenge.

Recent advancements in large language models (LLMs) have significantly accelerated progress in natural language processing (NLP), particularly in tasks such as clinical summarization and entity extraction~\citep{chen2024ed, aali2024hospital, ellershaw2024discharge}. In addition to the rapid development of general-purpose LLMs—such as \texttt{GPT-4}~\citep{achiam2023gpt}, \texttt{LLaMA 3}~\citep{grattafiori2024llama}, and \texttt{Gemma}~\citep{team2025gemma}—domain-specific models like \texttt{BioInstruct}~\citep{wang2023bioinstruct}, \texttt{MediSwift}~\citep{bhardwaj2024mediswift}, and \texttt{BioMedLM}~\citep{boag2024biomedlm} have been introduced to better capture the structure and semantics of biomedical corpora, improving the understanding of medical narratives. These types of models have demonstrated considerable promise in enhancing the accuracy and efficiency of clinical documentation and information extraction~\citep{chen2024ed, hu2024clinical}. However, they face critical limitations in real-world clinical settings. Specifically, unstructured clinical narratives are often verbose, redundant, and contain various types of nomenclature and jargon, increasing computational demands and obscuring essential information captured in the notes~\citep{hu2024longrecipe}. Reliance on fine-tuning and pretraining domain-specific LLMs further amplifies these challenges, as such processes require substantial computational resources and time~\citep{christophe2024med42, liu2024moe}. These constraints highlight the need for scalable, domain-aware methodologies that balance contextual fidelity with computational efficiency.
Model compression and optimization techniques such as pruning~\citep{ling2024slimgpt}, quantization~\citep{lang2024comprehensive}, and distillation~\citep{muralidharan2024compact} address some of these computational challenges by reducing the resource demands of deploying LLMs without significantly sacrificing their performance~\citep{zhu2024survey}. 
Such approaches are particularly relevant in healthcare, where the efficient processing of large volumes of data is crucial for timely and accurate patient care. Optimizing LLMs for clinical use can mitigate their high resource requirements, making them more suitable for integration into existing clinical workflows and systems.

Recent studies evaluating LLMs on clinical note summarization highlight significant limitations in fidelity and coherence. Models often generate factually incorrect or fabricated content (hallucinations), posing a major obstacle for clinical use~\citep{oeshy2024improving, fayyaz2024enabling, poulain2024aligning}. For example, a physician review of GPT-4-generated emergency department summaries found hallucinated details in 42\% of cases and omission of relevant clinical information in 47\%~\citep{williams2024evaluating}. Such omissions of key medications, diagnoses, or events are common, sometimes due to oversimplification of complex cases~\citep{lee2024prospects}. LLMs also struggle with temporal reasoning: they may misrepresent the chronology of care by focusing on outdated or irrelevant diagnoses (treating them as current) and not recognizing when earlier presumptive diagnoses were later ruled out~\citep{xiong2024large}. Even when factual coverage is adequate, the narrative quality can be suboptimal – generated summaries are often less concise and lack the realistic clinical tone or structured flow of human-written notes~\citep{van2024adapted}. Moreover, current LLMs face context length constraints, often requiring truncation or segmentation of long patient records; as a result, many evaluations use isolated note segments instead of full longitudinal records, risking loss of important context~\citep{ravaut2023context}. These limitations – hallucinations, omissions, poor handling of temporal context, and subpar coherence – underscore that while LLMs show promise in reducing documentation burden, they are not yet fully reliable for autonomous clinical summarization~\citep{wang2024ada,hager2024evaluating}.

To address these challenges, we propose \texttt{ConTextual}, a novel framework that integrates a context-preserving token filtering (CPTF) approach with a domain-specific knowledge graph (KG) to enhance clinical text summarization. CPTF leverages attention mechanisms to dynamically identify and retain semantically significant tokens, minimizing computational costs while preserving critical clinical information. To mitigate information loss from token filtering, the domain-specific KG encodes structured relationships among clinical entities, such as diagnoses, treatments, and outcomes. This integration ensures that the retained tokens are enriched with domain-relevant context, enabling the framework to maintain contextual fidelity in complex clinical scenarios.

\subsection*{Generalizable Insights about Machine Learning in the Context of Healthcare}
The proposed framework provides a scalable and efficient solution for processing verbose clinical narratives. By prioritizing clinically significant tokens and enriching them with structured knowledge, \texttt{ConTextual} achieves superior information retention, computational efficiency, and contextual depth. Our extensive evaluations demonstrate superior performance in summarization ability and reduction in latency and computational costs, making the framework particularly suited for complex and resource-constrained healthcare environments. 
Although tailored for clinical note summarization, the modular design of \texttt{ConTextual} supports broader applicability to other biomedical domains requiring efficient and domain-specific natural language understanding, such as medical literature review. In particular, our contributions are as follows:
\begin{itemize}
    \item We propose a context-preserving token filtering (CPTF) method that dynamically compresses unstructured clinical text by removing redundancy while retaining essential information.
    
    \item We construct a domain-specific knowledge graph (KG) and integrate it with the CPTF to form a structured and interpretable framework that enhances contextual fidelity during token selection.

    \item We improve the contextual input for retrieval-augmented generation (RAG), enabling more effective LLM-based reasoning and demonstrating improved performance across two clinical datasets.

    \item We validate the scalability and efficiency of \texttt{ConTextual} through extensive evaluations using metrics, including lexical, semantic, and LLM-based measures.
\end{itemize}

\section{Related Work}
\paragraph{Clinical Text Summarization}
Existing summarization models rely heavily on standard attention mechanisms, which scale quadratically with sequence length. For instance, \texttt{BioGPT}~\citep{luo2022biogpt} and \texttt{PubMedBERT}~\citep{gu2021domain} utilize biomedical corpora to refine performance, but their reliance on uncompressed token sequences results in inefficiencies for lengthy clinical notes. Models like \texttt{Flan-T5}~\citep{lyu2024automatic} have introduced instruction-tuned objectives to improve summarization; however, they may fail to address the redundancy of verbose clinical narratives, where attention mechanisms struggle to focus on critical contextual cues~\citep{hu2024longrecipe}. To mitigate these issues, recent work~\citep{han2024optimal} has proposed models such as \texttt{Pointer-GPT}, which replace standard attention mechanisms with a pointer network to enhance content retention during summarization. However, such models may still suffer from factual inconsistency, highlighting the ongoing need for summarization approaches that balance precision, coherence, and domain specificity.

\paragraph{Model Compression and Optimization}
While model compression techniques such as pruning~\citep{frantar2023sparsegpt}, quantization~\citep{dettmers2022gpt3}, and distillation~\citep{hinton2015distilling} effectively reduce model size and latency, they often degrade performance on domain-specific tasks due to loss of fine-grained contextual information~\citep{tinn2023fine}. For instance, pruning reduces model complexity by zeroing out low-magnitude weights, but in clinical NLP tasks, even small weights can encode critical semantic relationships~\citep{ma2023llm}. Similarly, knowledge distillation transfers knowledge from large to smaller models~\citep{ho2022large}, but these smaller models may lack the capacity to retain nuanced biomedical context~\citep{magister2022teaching}.
\paragraph{Token Filtering and Attention Mechanisms}
Token filtering methods(e.g.,\texttt{PoWER-BERT}), rely on attention scores to progressively prune less relevant tokens~\citep{goyal2020power}. However, these approaches operate in encoder-only architectures and are incompatible with the autoregressive decoding required by generative models typically employed in large language model applications. \texttt{Prunepert} introduced a differentiable perturbed top-k mechanism for token selection, but its reliance on stochastic perturbations increases variance in summarization outcomes. By contrast, \texttt{CPTF} operates natively within the multi-layered attention framework of modern open-source LLMs, dynamically weighting attention layers to compute token importance without introducing architectural modifications.

Moreover, long and heterogeneous texts significantly increase latency and computational overhead, which limits their scalability in resource-constrained clinical environments~\citep{he2025survey}. Token filtering methods aim to mitigate this by dynamically retaining contextually important tokens while discarding less relevant ones~\citep{lin2024rho, lou2024sparser}. While effective in reducing computational demands, these methods often result in partial information loss, particularly in complex, domain-specific scenarios such as healthcare. Additionally, they are not readily integrable with structured knowledge, such as knowledge graphs, which can compensate for the loss of contextual information~\citep{he2025survey, liu2024large}.

\paragraph{Knowledge Graph Integration in NLP}
KGs are extensively utilized for encoding structured relationships, offering enhanced contextualization and interpretability in NLP applications~\citep{peng2023knowledge, sharma2022knowledge}. In the biomedical domain, KGs like UMLS and SNOMED-CT have been employed for tasks such as entity linking and ontology-based query expansion~\citep{lu2025biomedical, arsenyan2024large, hu2023survey}. By encoding explicit relationships (e.g., between diseases, symptoms, and treatments), KGs can enhance a model’s contextual understanding and interpretability, helping align NLP outputs with established medical knowledge~\citep{he2025survey, liu2024large, piya2025advancing}. However, most prior approaches integrate KGs in a static fashion—treating the graph as a fixed resource—which means the knowledge base does not dynamically update or adapt to new data or task-specific needs, potentially limiting scalability and flexibility in fast-evolving clinical settings~\citep{review2024healthcare}. Recent work has begun exploring more dynamic KG integration strategies (e.g., continual graph updates or tailored subgraph retrieval) to improve adaptability~\citep{arsenyan2024large}, but static KGs remain invaluable as authoritative repositories of biomedical knowledge. In this context, our work \texttt{ConTextual} contributes a novel method that effectively leverages a static domain-specific biomedical KG within a clinical summarization framework, demonstrating that a fixed, curated graph can be harnessed to provide relevant structured context and significantly improve the factuality and clinical fidelity of LLM-generated summaries~\citep{lu2025biomedical}.

\section{Methods}

We propose \texttt{ConTextual}, a framework for clinical text summarization that addresses the challenges of long and verbose narratives through three key components: (1) Context-Preserving Token Filtering (CPTF), (2) Domain-specific KG integration, and (3) an LLM inference with retrieval-augmented generation (RAG). 

\begin{figure*}[tbp]
    \centering
    \includegraphics[width=\textwidth]{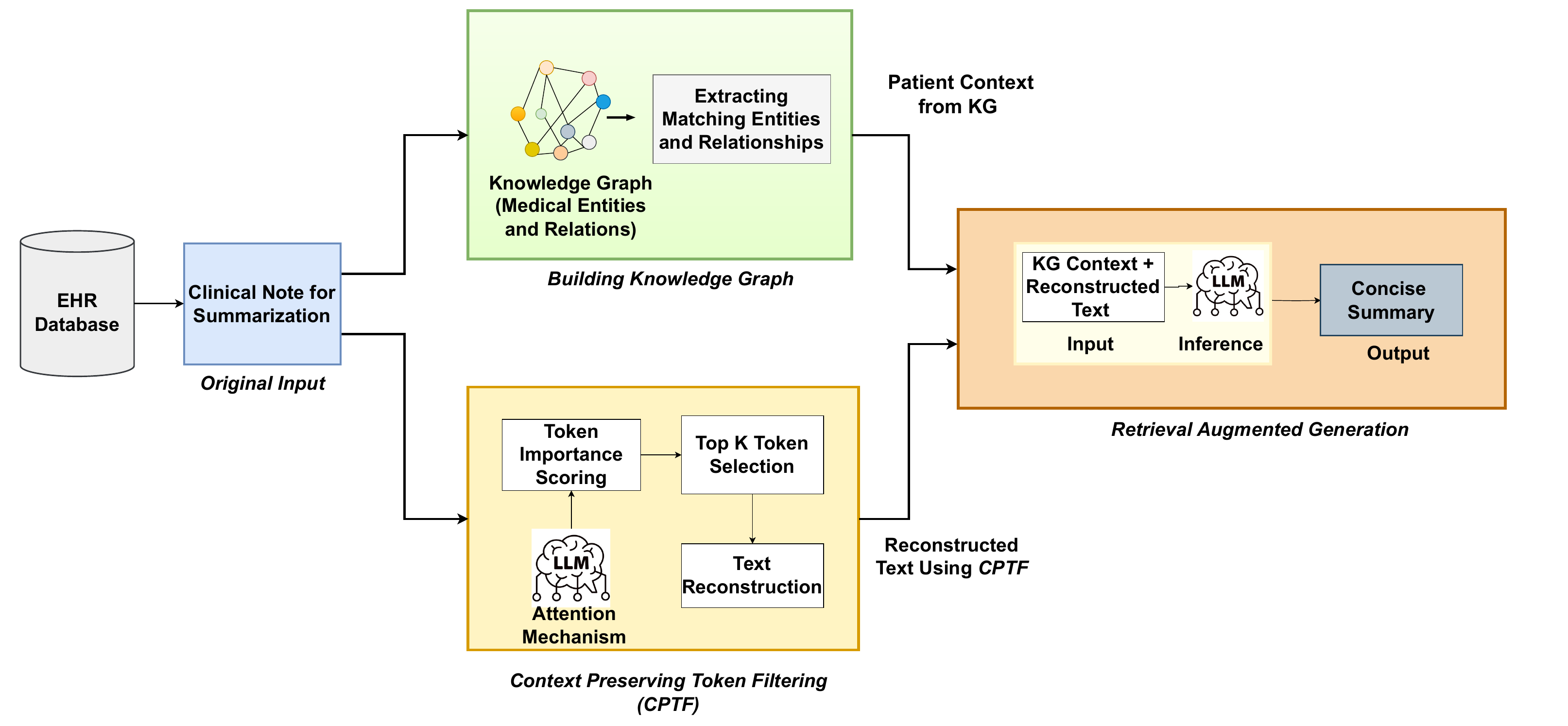}
    \caption{Overview of the \texttt{ConTextual} Framework for clinical text summarization. 
    }.
    \label{fig:system_overview}
\end{figure*}

CPTF dynamically reduces redundancy by filtering out tokens with low contextual significance based on the attention mechanisms of LLMs, ensuring that the retained input is computationally efficient and semantically rich. The KG integration component enriches this reduced input by embedding structured relationships among clinical entities, such as diagnoses, treatments, and medications, to provide domain-specific context. Finally, the RAG component retrieves additional relevant context from the KG during inference, integrating it into the summarization process to maintain accuracy and adaptability for specific queries. Figure~\ref{fig:system_overview} shows the overall structure of the framework and its three components. 

\subsection*{Problem Formulation} Prior to presenting the components of the proposed method in detail, we first present an overall description of the whole framework. 
Let $\mathcal{D} = \{d_1, ..., d_N\}$ denote the data elements representing a collection of clinical notes. Each note $d \in \mathcal{D}$ consists of an input sequence $d = \{t_1, ..., t_n\}$, where $t_i \in \mathcal{V}$ and $\mathcal{V}$ represents the vocabulary of tokens. Our objective is to generate a contextually-enhanced reduced representation $d_{reduced}$ that preserves critical medical information while minimizing the sequence length.

The text reduction task can be formalized as finding a mapping function $f \in \mathcal{F}$, where $\mathcal{F}$ represents a family of candidate functions. The optimization objective is:
\begin{equation}
\max_{f \in \mathcal{F}} \sum_{d \in \mathcal{D}} \text{sim}(d, f(d)),
\end{equation}
where $\text{sim}(d, f(d))$ shows the similarity between the original and reduced representations (measured by a measure like cosine). The reduced representation $f(d)$ satisfies the constraint:
\begin{equation}
|f(d)| = \lfloor r \cdot |d| \rfloor, \quad r \in (0,1]
\end{equation}
where $r$ denotes the \textit{retention ratio}, specifying the proportion of tokens retained in $d_{reduced}$.

To ensure coherence and clinical accuracy in the reconstructed and reduced text, LLM leverages knowledge from a reference medical knowledge graph, denoted as $G = (V, E, \mathcal{R})$, where $V$ is the set of nodes (e.g., medical entities), $E$ is the set of edges (relationships), and $\mathcal{R}$ represents the types of relationships. We define a context retrieval function $\eta: d \rightarrow 2^V$, where $2^V$ denotes the power set of $V$, i.e., the set of all subsets of $V$. This function retrieves a subset of relevant nodes from the KG based on the input sequence $d$.

The final summarization objective combines both the reduced text (through CPFT) and the enhanced context (through KG) as:
\begin{equation}
    s^* = \underset{s \in \mathcal{S}}{\operatorname{arg\,max}} P(s \mid [f(d); \eta(d)]; \theta),
\end{equation}
where $\mathcal{S}$ is the set of candidate summaries, $f(d)$ is the reduced text representation, $\eta(d)$ is the retrieved context, and $\theta$ denotes the model parameters. We now present the model and its three components in more detail. 


\subsection{Context Preserving Token Filtering (CPTF)}
The CPTF framework is illustrated in Figure~\ref{fig:cptf_overview}, and a detailed algorithm corresponding to this part is presented in Appendix~\ref{appendix:algorithm}. This framework processes clinical text sequences by leveraging multi-head attention mechanisms from LLM to compute token-level importance, thereby preserving semantic fidelity while optimizing computational efficiency.

We consider multiple attention heads ($B_1, B_2, \dots, B_n$) that independently compute token interactions, capturing diverse semantic aspects as depicted in Figure~\ref{fig:cptf_overview}. Each input token $i$ is first projected into three representations: \textit{query} ($q$), \textit{key} ($k$), and \textit{value} ($v$). The query ($q$) interacts with keys ($k$) across the sequence to compute attention weights, which are then applied to the values ($v$) to generate weighted representations of the input tokens.

The attention outputs from multiple heads are concatenated into a single vector ($B$), which aggregates critical semantic features across tokens. The extraction and selection of tokens are guided by their computed importance scores, allowing for the retention of the most contextually significant tokens. These tokens are then reconstructed into a reduced clinical narrative, maintaining the focus on preserving essential information with high semantic relevance.

\begin{figure*}[h]
    \centering
    \includegraphics[width=1\textwidth]{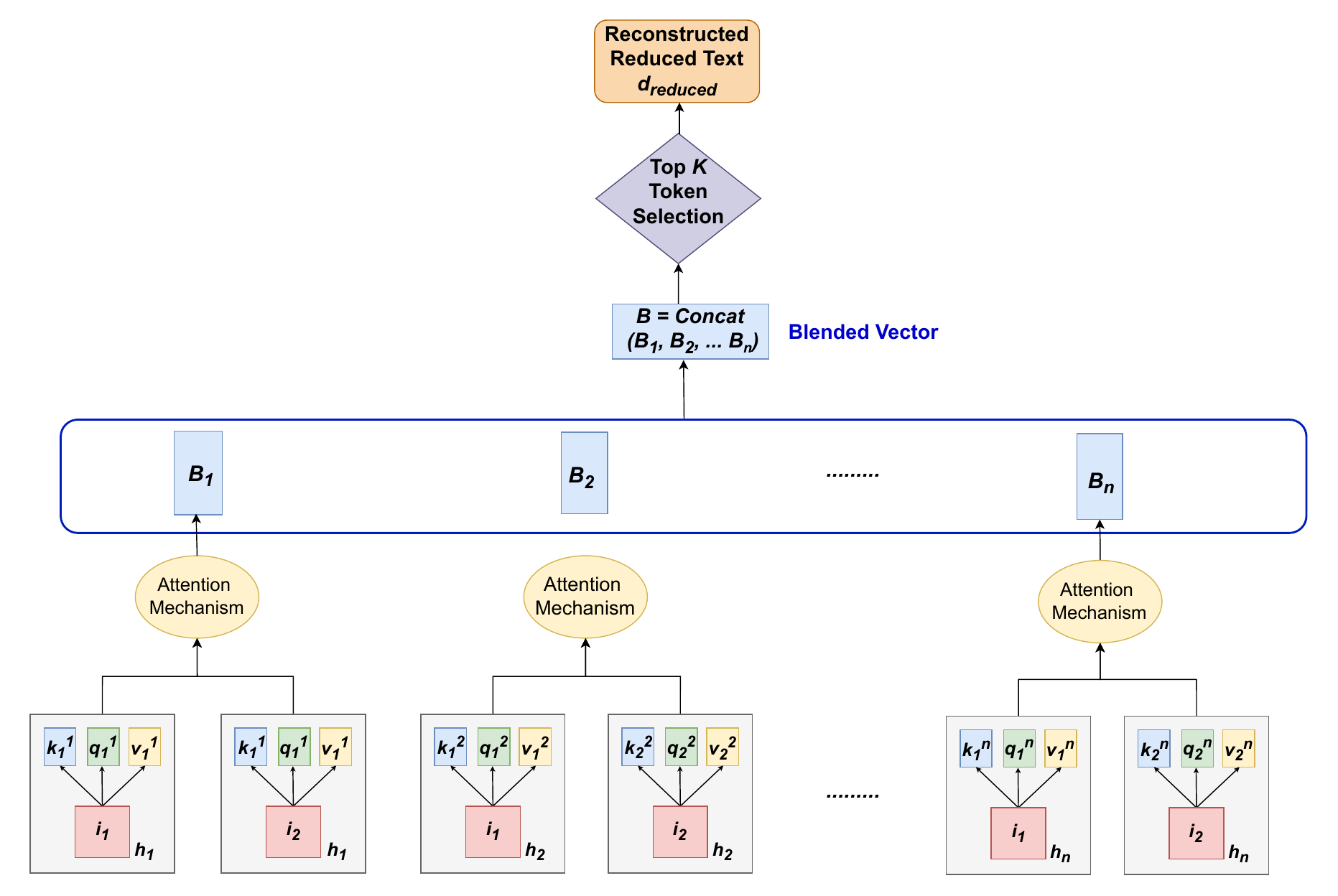}
    \caption{Overview of the Context Preserving Token Filtering algorithm.}
    \label{fig:cptf_overview}
\end{figure*}

\paragraph{Token Selection Mechanism}
Our methodology continues by extracting attention patterns from an LLM, wherein each input token is processed through transformer layers. Each layer $l \in \{1,..., L\}$ computes attention matrices $A_l^h \in \mathbb{R}^{n \times n}$, indicating token-to-token attention weights. The attention matrices from multiple heads are blended into a single matrix for each layer, calculated as:
\begin{equation}
    \bar{A}_l = \frac{1}{H} \sum_{h=1}^H A_l^h ,
\end{equation}
where $H$ is the total number of attention heads. This consolidated attention information helps in determining the hierarchical semantic structure across layers, ultimately influencing the selection of tokens that contribute most significantly to the clinical narrative's context and meaning. To capture the hierarchical semantic structure across layers, we compute layer weights as:
\begin{equation}
w_l = \alpha + (1 - \alpha) \frac{l}{L} , \label{eq:weighting_scheme}
\end{equation}
where $\alpha \in [0,1]$ is a tunable hyperparameter that balances the contribution of base-level features extracted from the lower layers with the more abstract features from higher layers. This balance aims to maintain a robust representation of both fundamental and complex features in the processed text, thereby preserving the integrity and richness of the clinical narrative.

\paragraph{Token Importance Scores and Selection}

The token importance score $I_i$ is computed for each token at position $i$ by aggregating the weighted attention patterns across all layers. This step captures both local syntactic relationships and global semantic dependencies:
\begin{equation}
   I_i = \sum_{l=1}^L w_l \cdot \frac{1}{n} \sum_{j=1}^n \bar{A}_l[i,j],
\end{equation}
where $w_l$ is the layer-specific weight, $n$ denotes sequence length, and $\bar{A}_l[i,j]$ represents the attention strength between tokens $t_i$ and $t_j$.
Using these scores, we select the top $k$ tokens by solving the constrained optimization problem:
\begin{equation}
    S^* = \underset{S \in \mathcal{S}}{\operatorname{arg\,max}} \sum_{i \in S} I_i ,
\end{equation}
subject to:
\begin{align}
   &|S| = k, \quad k = \lfloor r \cdot n \rfloor \\
   &\forall i,j \in S: i < j \implies \text{pos}(i) < \text{pos}(j) ,
\end{align}
where $r \in (0,1]$ is the retention ratio, and $\text{pos}(i)$ maps token position $i$ to its original sequence index.

\paragraph{Reconstruction of Reduced Narrative}

The final reduced sequence is reconstructed by mapping the selected token indices to their corresponding tokens and positional encodings:
\begin{equation}
   d_{reduced} = \{(t_i, \text{pos}(i)) \mid i \in S^*\}
\end{equation}

\subsection{Domain-specific Knowledge Graph Construction}
To enhance the understanding of the general medical context, we construct a reference KG \( G = (V, E, \mathcal{R}) \) using clinical records from the same cohort of patients for whom we analyze the clinical notes. The vertex set \( V \) comprises entity types representing the clinical domain hierarchy: \( V = V_d \cup V_m \cup V_t \), where \( V_d \), \( V_m \), and \( V_t \) correspond to diagnoses, medications, and treatments, respectively. These entities are interconnected through a set of edges \( E \subseteq V \times V \), which capture the general clinical relationships. The relationship types in the graph are defined as: \( \mathcal{R} = \{r_{dm}, r_{dt}, r_{mt}\} \), where \( r_{dm} \), \( r_{dt} \), and \( r_{mt} \)  represent the relationships between diagnoses and medications, diagnoses and treatments, and medications and treatments, respectively. These relationships are used to map out the complex interactions within the clinical data.

For a given clinical note \( d \), we utilize a retrieval function \( \eta: d \rightarrow V_p \), where \( V_p \subset V \) is the set of entities matching the unique patient identifier in the clinical note. Each entity \( e \in V_p \) represents a clinical concept linked to the patient. The contextual information retrieved from the KG is formalized as:
\begin{equation}
   C(d) = \{(e', r) \mid e \in V_p, (e, e', r) \in E\},
\end{equation}
where \( e' \) denotes an entity in the KG that shares a relationship \( r \) with \( e \).

This arrangement retrieves entities directly linked to the same patient identifier as in the clinical notes, ensuring a match of patient-specific entities and their relationships. The extracted context enriches the input by concatenating the reduced clinical note with the retrieved context, forming the final enriched input for downstream processing:
\begin{equation}
   \hat{d} = d_{\text{reduced}} \oplus C(d),
\end{equation}
where, \( \oplus \) denotes concatenation. This approach ensures that the enriched representation \( \hat{d} \) retains key patient-specific details while incorporating domain knowledge from the knowledge graph.

\subsection{Summarization with RAG}
The summarization process leverages the patient-specific reduced text from CPTF $d_{\text{reduced}}$ and dynamically retrieved KG context $C(d)$ using a RAG approach. 
The summarization objective aims to generate the most clinically relevant and coherent summary \( s^* \) from a set of possible summaries \( \mathcal{S} \). This optimal summary is selected by the model as:
\begin{equation}
   s^* = \underset{s \in \mathcal{S}}{\operatorname{arg\,max}} P(s \mid \hat{d}; \theta),
\end{equation}
where \( \theta \) denotes the parameters of the model (i.e., the summary generator LLM). This approach conditions the generation of summaries on both the extracted entities and their associated knowledge from the KG, allowing the LLM to refine its understanding of patient-specific narratives and improve factual consistency and medical coherence. Since entity retrieval is performed via overlapping patient identifiers, the retrieved context remains directly relevant to the clinical note, ensuring adaptability across diverse cases. By dynamically incorporating structured medical knowledge from the KG, this approach enables LLMs to generate clinically coherent and factually consistent summaries while maintaining computational efficiency.

\section{Experiments}

\subsection{Experimental Setup}
We implement both the CPTF module and the primary summary generation component using the instruction-tuned LLaMA 3.2 1B model~\citep{llama3_1b}. All experiments are conducted with a fixed generation budget of 200 tokens and a decoding temperature of 0.7 to ensure consistency and comparability across runs.
\paragraph{Data}
For the summarization task, we utilized the MIMIC-IV-Ext-BHC dataset~\citep{aali2024mimic}, derived from the MIMIC-IV-Note database, consisting of 270,033 clinical notes with corresponding brief hospital course (BHC) summaries. The preprocessing involved standardizing the structure of the note, cleaning the formatting, and normalizing the length of the token to an average of 2,267 tokens per note. The resulting curated dataset provides a structured resource for clinical text summarization research~\citep{aali2024benchmark}. 

The second dataset we incorporated comprises 1,473 patient-doctor conversations from the FigShare~\citep{singh2011figshare} and MTS-Dialog~\citep{abacha2023empirical} collections, specifically designed for generating clinical summaries. These conversations have been annotated to create structured SOAP (Subjective, Objective, Assessment, and Plan) summaries, available on Hugging Face datasets~\citep{neupane2024soap}.

\paragraph{Prompt Design}
We use a few-shot design to provide the language model with structured input-output examples that establish a consistent format for clinical summarization. Few-shot prompting outperformed zero-shot and one-shot strategies across key evaluation metrics,  as shown in Table~\ref{tab:prompting-strategies} (Appendix). These observed improvements indicate that providing the model with representative clinical examples helps constrain its output format, enhancing factual coherence and reducing hallucinations. Additionally, few-shot learning ensures that the model remains aligned with clinical terminology and structured reporting conventions, addressing concerns about variability in generated summaries.

The offered examples in the few-shot design align structured input tags—such as \texttt{<SEX>}, \texttt{<SERVICE>}, \texttt{<CHIEF COMPLAINT>}, and \texttt{<HISTORY OF PRESENT ILLNESS>}—with corresponding target summaries, capturing linguistic patterns and contextual nuances prevalent in clinical narratives. A description of the few-shot prompt design, including illustrative examples, is provided in Table~\ref{tab:prompt-example}.

\paragraph{Baselines}
We selected a range of models, chosen for their diverse approach to handling the challenges of clinical text summarization:

\begin{compactitem}
    \item \texttt{Longformer} by~\citet{beltagy2020longformer} utilizes a sparse attention mechanism to handle long documents by extending attention spans up to 16K tokens, particularly suited for detailed clinical narratives.
    \item \texttt{BioBART} by~\citet{yuan2022biobart} is specifically pre-trained on biomedical corpora and fine-tuned for medical summarization.
    \item \texttt{T5-Large} by~\citet{raffel2020exploring} is a general-purpose sequence-to-sequence model that excels in diverse text-to-text tasks, testing the adaptability of transformers in specialized domains.
    \item \texttt{Flan-T5} by ~\citet{chung2022scaling} extends \texttt{T5} with instruction tuning, aiming to improve the model's ability to learn from descriptive tasks and generalizing across various NLP applications.
    \item \texttt{BioGPT} by~\citet{BioGpt} offers an adaptation of the GPT architecture tailored to understand and generate biomedical text, focusing on maintaining clinical accuracy and relevance.
    \item \texttt{Gemma3-Instruct(1B)} by Google~\citep{team2025gemma} is a open-source general-purpose LLM.
    \item \texttt{Mistral-7B-Instruct} by Mistral AI ~\citep{jiang2023mistral} is the second open-source general-purpose LLM we use.
\end{compactitem}

\paragraph{Evaluation Metrics}


We use the following statistical metrics to evaluate the accuracy and relevance of the generated summaries.

\begin{compactitem}
\item \textbf{BLEU}~\citep{papineni2002bleu}: Measures n-gram precision to quantify lexical overlap between generated and reference summaries.

\item \textbf{ROUGE-L}~\citep{lin2004rouge}: Captures the longest common subsequence between generated and reference summaries, emphasizing recall and precision while reducing redundancy.

\item \textbf{BERT-Score}~\citep{zhang2019bertscore}: Computes semantic similarity using contextual embeddings, ensuring accurate representation of clinical meaning.
\end{compactitem}

We evaluate using BLEU-1 (B-1), BLEU-2 (B-2), and ROUGE-L (R-L) for surface-level fidelity, and precision (P), recall (R), and F1-score derived from BERTScore to quantify semantic alignment with gold-standard references.

To complement statistical evaluation metrics and better assess the clinical fidelity and coherence of generated summaries, we also employ an `LLM as a judge' framework. We use an instruction-tuned \texttt{Gemma 3.1B} model~\citep{team2025gemma} to evaluate the generated summaries relative to their corresponding reference along three critical axes: Main Idea Retention, Coherence, and Factual Consistency. Evaluators—instantiated through LLM prompting—were instructed to assign a score from 1 (poor) to 5 (excellent) for each criterion, guided by a structured evaluation prompt.

Additionally, we use two metrics to study the scalability and resource optimization of the proposed method.

\begin{compactitem}
    \item \textbf{Throughput}~\citep{vaswani2017attention}: Calculates summaries generated per second, showcasing scalability for large datasets.

\item\textbf{Latency}~\citep{narang2021transformer}: Evaluates the time taken to generate a single summary, reflecting the computational cost and efficiency of different prompting strategies.
\end{compactitem}

\subsection{Results}
\paragraph{Ablation Analysis} 
Table~\ref{tab:model-performance-comparison} reports the performance of three model configurations evaluated on the MIMIC-BHC and SOAP datasets. We compare the full model against two systematically reduced variants: (i) a baseline LLM without Context-Preserving Token Filtering (Vanilla LLM) and (ii) an intermediate variant that incorporates CPTF but excludes Knowledge Graph augmentation (Vanilla LLM + CPTF). The consistent performance degradation across both datasets—particularly in ROUGE-L and BERT-F1—underscores the complementary contributions of CPTF and KG integration to the overall effectiveness of the proposed framework.

\begin{table*}[h!]
    \centering
    \caption{\textbf{Performance Comparison of Clinical Summarization Models on MIMIC-BHC and SOAP Datasets.}  BERT-P, BERT-R, and BERT-F1 refer to precision, recall, and F1 score using BERT embeddings.}
    \label{tab:model-performance-comparison}
    \footnotesize
    \resizebox{0.99\textwidth}{!}{
    \renewcommand{\arraystretch}{1.1}
    \begin{tabular}{@{}p{1.6cm}|l|cccccc@{}}
        \toprule
        \textbf{Dataset} & \textbf{Model} & \textbf{BLEU-1 (↑)} & \textbf{BLEU-2 (↑)} & \textbf{ROUGE-L (↑)} & \textbf{BERT-P (↑)} & \textbf{BERT-R (↑)} & \textbf{BERT-F1 (↑)} \\
        \midrule
        \multirow{3}{*}{\parbox{1.6cm}{\textbf{MIMIC-\\BHC}}} 
        & \textbf{\texttt{LLaMA 3.2}}  & $4.52 \pm 6.1$ & $1.79 \pm 2.6$ & $8.85 \pm 3.5$ & $\mathbf{83.02} \pm 2.0$ & $78.64 \pm 2.2$ & $80.77 \pm 1.7$ \\
        & \textbf{\texttt{LLaMA 3.2 +CPTF}} & $5.99 \pm 5.5$ & $1.82 \pm 2.0$ & $7.08 \pm 2.9$ & $81.82 \pm 2.4$ & $80.12 \pm 2.1$ & $80.97 \pm 1.8$ \\
        & \textbf{\texttt{ConTextual}} & $\mathbf{9.06} \pm 6.1$ & $\mathbf{3.35} \pm 2.6$ & $\mathbf{9.98} \pm 3.5$ & $82.72 \pm 2.0$ & $\mathbf{80.32} \pm 2.2$ & $\mathbf{81.48} \pm 1.7$ \\
        \midrule
        \multirow{3}{*}{\parbox{1.6cm}{\textbf{SOAP\\Summary}}} 
        & \textbf{\texttt{LLaMA 3.2}} & $4.13 \pm 5.5$ & $2.22 \pm 3.7$ & $8.16 \pm 6.5$ & $82.87 \pm 2.8$ & $82.93 \pm 3.0$ & $82.90 \pm 2.8$ \\
        & \textbf{\texttt{LLaMA 3.2 + CPTF}} & $9.29 \pm 7.3$ & $3.98 \pm 4.1$ & $8.75 \pm 5.5$ & $82.58 \pm 3.2$ & $82.38 \pm 2.8$ & $82.45 \pm 2.6$ \\
        & \textbf{\texttt{ConTextual}} & $\mathbf{11.55} \pm 5.5$ & $\mathbf{6.09} \pm 4.2$ & $\mathbf{10.70} \pm 4.5$ & $\mathbf{83.51} \pm 1.6$ & $\mathbf{83.70} \pm 3.0$ & $\mathbf{83.60} \pm 2.2$ \\
        \bottomrule
    \end{tabular}
    }
\end{table*}

On MIMIC-BHC, ConTextual achieves a BERT-F1 of $81.48 \pm 1.7$, outperforming LLaMA 3.2 ($80.77 \pm 1.7$) and CPTF-enhanced LLaMA 3.2 ($80.97 \pm 1.8$). A similar pattern is observed on the SOAP dataset, where ConTextual attains the highest BERT-F1 of $83.60\pm 2.2$ compared to $82.90 \pm 2.8$ and $82.45 \pm 2.6$ from the respective baselines. Improvements are also evident in lexical metrics: on SOAP, ConTextual increases BLEU-1 from 4.13 to $\mathbf{11.55}$ and BLEU-2 from 2.22 to $\mathbf{6.09}$, indicating enhanced surface-level coherence and informativeness.
We also explore the effects of temperature scaling and token limit adjustments and present the results in Appendix~\ref{appendix:cptf}.


We also use the  LLM as a judge evaluator to score each output according to defined criteria. Table~\ref{tab:llm-judge-eval} presents the mean and standard deviation of these scores across datasets and models. Each column corresponds to one of the three evaluation criteria, with the final column (Avg.) representing the average of the three scores per instance. 
\begin{table*}[htbp]
    \centering
    \caption{\textbf{LLM-as-a-Judge Evaluation.} Scores have a 1-5 scale.}
    \label{tab:llm-judge-eval}
    \footnotesize
    \resizebox{1\textwidth}{!}{
    \renewcommand{\arraystretch}{1.0}
    \begin{tabular}{@{}l|l|cccc@{}}
        \toprule
        \textbf{Dataset} & \textbf{Model} & \textbf{Main Ideas (↑)} & \textbf{Coherence (↑)} & \textbf{Factuality (↑)} & \textbf{Average Score (↑)} \\
        \midrule
        \multirow{3}{*}{\parbox{1.4cm}{\textbf{MIMIC-BHC}}} 
        & \textbf{\texttt{LLaMA 3.2}} & \textbf{4.56 $\pm$ 1.33} & 3.78 $\pm$ 0.67 & 3.89 $\pm$ 1.17 & 4.07 $\pm$ 0.95 \\
        & \textbf{\texttt{LLaMA 3.2 + CPTF}} & 3.93 $\pm$ 1.33 & 3.38 $\pm$ 0.77 & 3.77 $\pm$ 1.09 & 3.73 $\pm$ 0.87 \\
        & \textbf{\texttt{ConTextual}} & 4.45 $\pm$ 1.21 & \textbf{3.82 $\pm$ 0.75} & \textbf{4.55 $\pm$ 0.82} & \textbf{4.27 $\pm$ 0.55} \\
        \midrule
        \multirow{3}{*}{\parbox{1.4cm}{\textbf{SOAP Summary}}} 
        & \textbf{\texttt{LLaMA 3.2}} & 4.39 $\pm$ 0.85 & 3.71 $\pm$ 0.85 & 4.06 $\pm$ 1.03 & 4.09 $\pm$ 0.69 \\
        & \textbf{\texttt{LLaMA 3.2 + CPTF}} & 4.70 $\pm$ 0.66 & 3.80 $\pm$ 0.52 & 3.84 $\pm$ 1.17 & 4.13 $\pm$ 0.60 \\
        & \textbf{\texttt{ConTextual}} & \textbf{5.00 $\pm$ 0.00} & \textbf{4.50 $\pm$ 0.55} & \textbf{4.67 $\pm$ 0.52} & \textbf{4.72 $\pm$ 0.25} \\
        \bottomrule
    \end{tabular}
    }
\end{table*}
\texttt{ConTextual} stands out as the winner model across the evaluation dimensions, consistently achieving the highest scores in most criteria. Its performance advantage is particularly evident in the SOAP dataset, where it attains near-ceiling ratings with minimal variance. These results position \texttt{ConTextual} as the most effective approach among those evaluated, highlighting the value of integrating context-preserving token filtering with structured knowledge representations in clinically grounded summarization tasks.

\paragraph{Comparison with Baselines}
\begin{table*}[hbtp!]
    \centering
    \caption{Performance Comparison with Baseline Models Across Datasets.}
    \label{tab:baseline-models}
    \footnotesize
    \setlength{\tabcolsep}{4pt}
    \renewcommand{\arraystretch}{0.9}
    \begin{tabular}{@{}ll|cccc@{}}
        \toprule
        \textbf{Dataset} & \textbf{Model} & \textbf{BLEU-1 (↑)} & \textbf{BLEU-2 (↑)} & \textbf{ROUGE-L (↑)} & \textbf{BERT F1 (↑)} \\
        \midrule
        \multirow{8}{*}{MIMIC-BHC} 
        & Longformer                & $2.76 \pm 5.0$  & $0.75 \pm 2.0$  & $3.10 \pm 3.0$  & $74.70 \pm 1.5$  \\ 
        & BioBART                   & $6.88 \pm 5.2$  & $2.05 \pm 2.1$  & $8.00 \pm 3.2$  & $78.10 \pm 1.6$  \\ 
        & T5-Large                  & $4.95 \pm 5.3$  & $1.41 \pm 2.2$  & $7.96 \pm 3.3$  & $79.84 \pm 1.6$  \\
        & Flan-T5                   & $10.52 \pm 5.8$ & $2.54 \pm 2.3$  & $9.90 \pm 3.4$  & $77.91 \pm 1.6$  \\ 
        & BioGPT                    & $6.15 \pm 5.4$  & $6.17 \pm 2.5$  & $7.47 \pm 3.2$  & $77.83 \pm 1.6$  \\
        & Gemma3-Instruct(1B)       & $7.89 \pm 5.6$  & $3.20 \pm 2.4$  & $9.85 \pm 3.4$  & $79.78 \pm 1.6$  \\
        & Mistral-7B-Instruct       & $4.43 \pm 5.2$  & $2.09 \pm 2.2$  & $9.87 \pm 3.4$  & $80.71 \pm 1.7$  \\
        & \textbf{\begin{tabular}[t]{@{}l@{}}ConTextual (Ours)\\\end{tabular}} 
                                    & $\mathbf{12.63} \pm 6.1$ & $\mathbf{4.65} \pm 2.6$ & $\mathbf{11.04} \pm 3.5$ & $\mathbf{81.37} \pm 1.7$ \\
        \midrule
        \multirow{8}{*}{SOAP Summary} 
        & Longformer                & $2.18 \pm 3.6$ & $1.19 \pm 2.6$ & $6.76 \pm 5.8$ & $75.00 \pm 2.9$ \\ 
        & BioBART                   & $5.27 \pm 4.2$ & $1.83 \pm 2.8$ & $7.41 \pm 4.9$ & $77.32 \pm 2.5$ \\ 
        & T5-Large                  & $3.86 \pm 3.9$ & $1.24 \pm 2.4$ & $7.52 \pm 5.2$ & $78.46 \pm 2.7$ \\
        & Flan-T5                   & $8.35 \pm 5.1$ & $2.17 \pm 2.9$ & $8.91 \pm 5.0$ & $77.05 \pm 2.8$ \\ 
        & BioGPT                    & $5.62 \pm 4.8$ & $5.11 \pm 3.7$ & $7.08 \pm 4.7$ & $76.93 \pm 2.6$ \\
        & Gemma3-Instruct(1B)       & $7.25 \pm 5.3$ & $2.84 \pm 3.2$ & $8.97 \pm 5.3$ & $79.25 \pm 2.5$ \\
        & Mistral-7B-Instruct       & $4.08 \pm 4.4$ & $1.85 \pm 2.7$ & $9.34 \pm 5.6$ & $81.18 \pm 2.4$ \\
        & \textbf{\begin{tabular}[t]{@{}l@{}}ConTextual (Ours)\\\end{tabular}} 
                                    & $\mathbf{11.55} \pm 5.5$ & $\mathbf{6.09} \pm 4.2$ & $\mathbf{10.70} \pm 4.5$ & $\mathbf{83.60} \pm 2.2$ \\
        \bottomrule
    \end{tabular}
\end{table*}
As shown in Table~\ref{tab:baseline-models}, \texttt{ConTextual} consistently outperforms all baselines across both datasets and evaluation metrics. The improvements in BLEU-1 and BLEU-2 reflect superior lexical overlap with reference summaries, while higher ROUGE-L and BERT F1 scores suggest stronger structural alignment and semantic preservation. These gains are particularly pronounced on the MIMIC-BHC dataset, highlighting our model’s effectiveness in handling longer and more complex clinical narratives. Notably, even when compared with instruction-tuned models such as Mistral-7B and Gemma3, \texttt{ConTextual} delivers higher performance, underscoring the value of structured retrieval and token filtering in medical summarization tasks.

\begin{table*}[ht]
    \centering
    \caption{\textbf{Efficiency Metrics Comparison Across Models.}}
    \label{tab:efficiency-comparison}
    \footnotesize
    \setlength{\tabcolsep}{5pt}
    \renewcommand{\arraystretch}{1.1}
    \begin{tabular}{@{}p{1.6cm}|l|cc@{}}
        \toprule
        \textbf{Dataset} & \textbf{Model} & \textbf{Throughput (↑)} & \textbf{Latency (↓)} \\
        \midrule
        \multirow{3}{*}{\parbox{1.6cm}{\textbf{MIMIC-\\BHC}}} 
        & \texttt{LLaMA 3.2} & $36.72 \pm 2.44$ & $\mathbf{3.61 \pm 1.32}$ \\
        & \texttt{LLaMA 3.2 + CPTF} & $139.10 \pm 10.04$ & $12.38 \pm 1.25$ \\
        & \texttt{ConTextual} & $\mathbf{142.87 \pm 16.94}$ & $14.29 \pm 2.17$ \\
        \midrule
        \multirow{3}{*}{\parbox{1.6cm}{\textbf{SOAP\\Summary}}} 
        & \texttt{LLaMA 3.2} & $28.35 \pm 36.26$ & $16.15 \pm 10.69$ \\
        & \texttt{LLaMA 3.2 + CPTF} & $108.43 \pm 12.75$ & $\mathbf{14.22 \pm 2.73}$ \\
        & \texttt{ConTextual} & $\mathbf{116.82 \pm 19.64}$ & $15.78 \pm 3.42$ \\
        \bottomrule
    \end{tabular}
\end{table*}

\paragraph{Computational Efficiency} Table~\ref{tab:efficiency-comparison} reports the efficiency characteristics of the proposed models and the relationship between computational performance and knowledge integration. Notably, both CPTF and \texttt{ConTextual} improve throughput compared to the base \texttt{LLaMA} 3.2 model, suggesting that context-aware token filtering enhances generation efficiency by reducing irrelevant token processing. The latency increases in the case of the MIMIC-BHC dataset, possibly due to the additional steps introduced by structured retrieval and filtering. This may reflect the trade-off between performance and computational efficiency, where modest increases in processing time are offset by substantial gains in output quality and overall generation efficiency.

\section{Discussion} 
This work introduces \texttt{ConTextual}, a structured framework for clinical text summarization that combines context-preserving token filtering (CPTF) with domain-specific knowledge graphs (KGs). CPTF dynamically reduces textual redundancy while preserving essential clinical information, and KG integration ensures that token selection aligns with structured domain knowledge. By improving the quality of inputs in a RAG structure, \texttt{ConTextual} enables more accurate and semantically grounded LLM-based reasoning. We validate the framework across two clinical datasets, demonstrating improvements in both generation quality and system efficiency, as measured by BLEU, ROUGE, BERTScore, LLM-based scores, latency, and throughput. Notably, it achieves up to a 1.5x improvement in BLEU-1 and a 31\% increase in ROUGE-L on SOAP summaries while delivering the highest BERTScore-F1 across all settings.  Beyond summarization, this framework has broader applicability in real-world clinical environments: alleviating documentation burden for providers, streamlining cohort identification for clinical trial recruitment, and enabling small to mid-sized healthcare organizations to deploy high-quality language models under constrained computational budgets.

\paragraph{Limitations} While \texttt{ConTextual} advances clinical summarization, it is constrained by its reliance on a static, domain-specific knowledge graph. This may restrict its generalizability to broader or evolving clinical domains, particularly in the context of rare conditions or emerging practices. Additionally, the framework assumes consistent quality and structure in clinical documentation, which may limit its robustness when applied to noisy, incomplete, or institution-specific records. Our preliminary analysis showed the value of developing a KG specific to the targeted domain (as determined by the input data). Future work can incorporate strategies for leveraging publicly available KGs to enhance adaptability. We also plan to support dynamic knowledge graph construction, expand entity coverage, and implement mechanisms for handling variability in input quality. These enhancements can further improve generalizability and resilience across diverse healthcare settings.

\begin{ack}
Our study was supported by the NIH award U54-GM104941 and P20-GM103446, as well as computing credits from Amazon Web Services (AWS).
\end{ack}


\bibliography{MLHC}
\bibliographystyle{unsrtnat}

\newpage
\appendix
\section{Context-Preserving Token Filtering Algorithm}
\label{appendix:algorithm}
We introduce an algorithm—\textit{Context-Preserving Token Filtering (CPTF)}—designed to retain semantically important tokens from an input sequence while minimizing overall length. By leveraging internal attention dynamics from a multi-layer transformer model, CPTF computes layer-weighted token importance scores to identify and preserve tokens that are critical for maintaining contextual coherence and clinical accuracy. While our implementation uses the instruction-tuned \texttt{LLaMA 3.2 1B}, the method is model-agnostic and applicable to a range of transformer-based language models. Its computational efficiency and minimal architectural assumptions make it well-suited for deployment in low-resource or compute-limited settings.
\begin{algorithm}[h]
\SetAlgoLined
\caption{Context-Preserving Token Filtering (CPTF)}
\label{appendix:cptf}
\DontPrintSemicolon
\KwIn{Text sequence $x$ (string), language model $M$ with $L$ layers and $H$ attention heads, tokenizer $T$, retention ratio $r \in (0,1]$ (float), base weight $\alpha \in [0,1]$ (float)}
\KwOut{Reduced sequence (string) retaining the most informative tokens}

\textbf{Step 1: Tokenization and Initialization}\;
$tokens \gets T.\text{encode}(x)$ \tcp*{Convert text to tokens}
$n \gets |tokens|$ \tcp*{Determine total number of tokens}
$k \gets \lfloor n \cdot r \rfloor$ \tcp*{Determine number of tokens to retain}
$I \gets [0] * n$ \tcp*{Initialize importance scores array}

\textbf{Step 2: Calculate Token Importance}\;
\For{$l = 1$ \KwTo $L$}{
    $w_l \gets \alpha + (1 - \alpha) \cdot \frac{l}{L}$ \tcp*{Layer-specific weight}
    $A_l \gets M.\text{attention}(tokens, l)$ \tcp*{Compute attention for layer $l$}
    $\bar{A}_l \gets \frac{1}{H} \sum_{h=1}^H A_{l,h}$ \tcp*{Average over all heads}
    \For{$i = 1$ \KwTo $n$}{
        $I[i] \gets I[i] + w_l \cdot \frac{1}{n} \sum_{j=1}^n \bar{A}_l[i, j]$ \tcp*{Update importance score}
    }
}

\textbf{Step 3: Token Selection and Reconstruction}\;
$S \gets \text{argsort}(-I)[:k]$ \tcp*{Select indices of top-$k$ important tokens}
$tokens_{\text{reduced}} \gets [tokens[i] for i in \text{sorted}(S)]$ \tcp*{Retrieve and sort tokens}
\Return $T.\text{decode}(tokens_{\text{reduced}})$ \tcp*{Reconstruct reduced sequence}
\end{algorithm}

\section{Knowledge Graph Construction}
\label{appendix:kg}
Traditional retrieval-augmented generation (RAG) models face limitations in synthesizing information from diverse sources, particularly when understanding requires identifying shared attributes or underlying semantic relationships~\cite{edge2024local}. To address these limitations, we constructed a domain-specific knowledge graph (KG) that encapsulates critical aspects of patient-level clinical data. The KG represents key entities, including \textit{Problems}, \textit{Treatments}, \textit{Tests}, and \textit{Patients}, along with their relationships, providing a structured and queryable representation of clinical interactions.

The constructed KG consists of 7,095 nodes distributed as follows: \textbf{Patients} (100 nodes), \textbf{Problems} (3,841 nodes), \textbf{Treatments} (1,686 nodes), and \textbf{Tests} (1,468 nodes). These entities are interconnected through 11,443 relationships categorized into three primary types: \texttt{HAS\_PROBLEM} (6,760 edges), \texttt{UNDERWENT\_TEST} (5,469 edges), and \texttt{WAS\_TREATED\_WITH} (3,214 edges). These relationships encode clinically meaningful associations, such as diagnoses associated with patients, tests performed, and treatments administered.

Entity extraction was performed using a domain-specific named entity recognition (NER) pipeline built with the \texttt{samrawal/bert-base-uncased\_clinical-ner} model. To process long clinical narratives, the pipeline segmented text into overlapping chunks, adhering to the model's token limits while preserving entity coherence. Extracted entities were grouped into the predefined categories and linked to patients, forming the basis for graph construction. Overlapping patient nodes denote cases where multiple diagnoses, tests, and treatments are associated with a single individual, effectively capturing the multi-relational structure inherent in clinical datasets.

The graph structure supports dynamic traversal to extract contextually relevant subgraphs for specific queries or tasks. For example, when responding to a patient-centric query, the KG enables the efficient extraction of related diagnoses, tests, and treatments, leveraging its multi-relational structure to ensure specificity and precision. This capability facilitates integration with downstream tasks, such as retrieval-augmented generation (RAG), contextual language modeling, and predictive analytics.

\begin{figure*}[h!]
    \centering
    \includegraphics[width=0.9\textwidth]{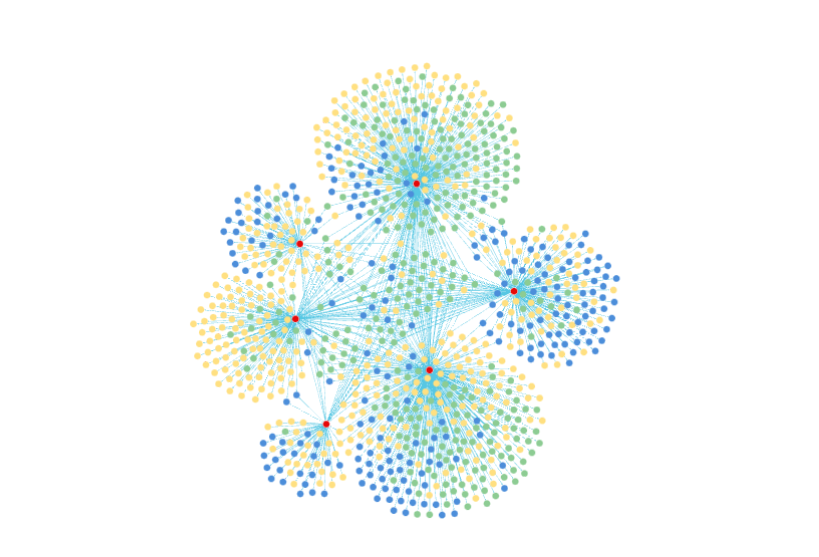}
    \caption{Visualization of the clinical knowledge graph constructed from unstructured EHR data. The graph comprises 1,000 nodes, categorized into \textbf{Patients} (\textcolor{red}{red}), \textbf{Problems} (\textcolor{yellow}{yellow}), \textbf{Tests} (\textcolor{green}{green}), and \textbf{Treatments} (\textcolor{blue}{blue}). Relationships between nodes (\textbf{1,187 edges}) include \textit{HAS\_PROBLEM}, \textit{UNDERWENT\_TEST}, and \textit{WAS\_TREATED\_WITH}, encoding critical clinical entity interactions. This structured representation facilitates interpretable and domain-aware contextualization for downstream tasks, such as summarization and retrieval.}
    \label{fig:knowledge_graph_visualization}
\end{figure*}

Figure~\ref{fig:knowledge_graph_visualization} illustrates a subset of the constructed KG, showing the relationships between patients, problems, treatments, and tests. By embedding clinical data into a graph structure, the KG provides a scalable and interpretable framework for contextualizing patient information and supporting advanced machine learning applications in the healthcare domain.
In addition to the MIMIC-IV dataset, we applied the same knowledge graph construction pipeline to a structured SOAP-format clinical summary dataset. This yielded a domain-specific graph aligned with the Subjective, Objective, Assessment, and Plan sections. While the core methodology remained consistent, minor adaptations were introduced to account for the structural characteristics of patient–provider dialogue. The resulting graph was similarly integrated into our retrieval framework to evaluate the generalizability of our approach across diverse clinical documentation formats.

\section{Layer Weighting Strategy}
\label{appendix:layer}
\begin{figure}[h!]
    \centering
    \includegraphics[width=0.5\linewidth]{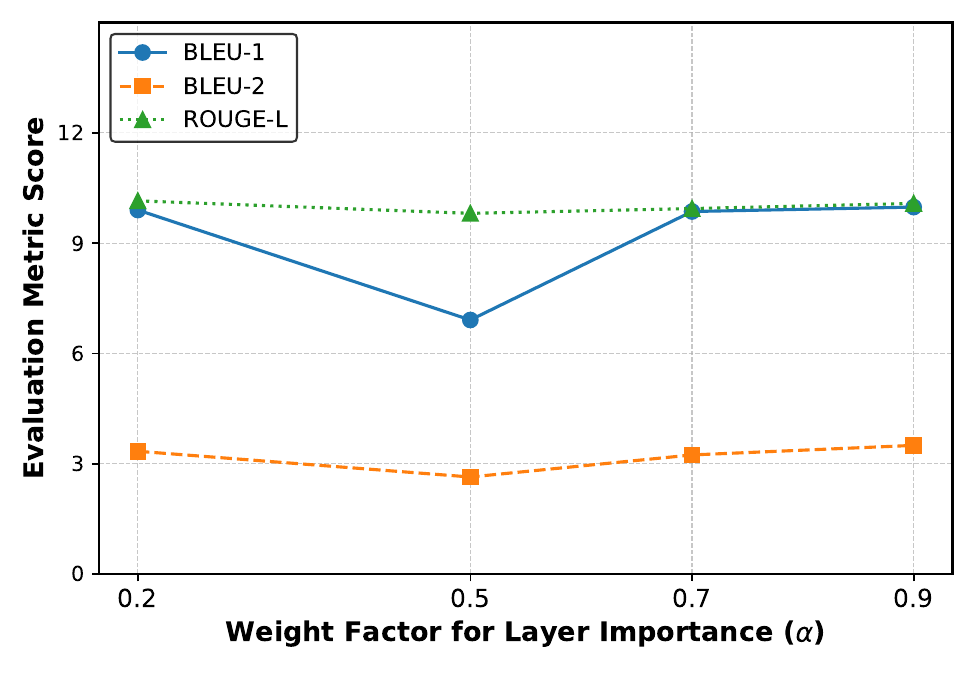}
    \caption{Comparing results with Weight factor for layer importance}
    \label{fig:alpha_performance}
\end{figure}

The choice of \( \alpha = 0.5 \) for our experiments was driven by its role in balancing the contributions of features from different transformer layers, which is pivotal for achieving a robust performance across various metrics. This balance ensures an effective integration of nuanced, deep contextual information from lower layers with more immediate, surface-level features from higher layers, thus preventing the model from overfitting to syntactic structures at the expense of semantic coherence. Such equilibrium is essential for the application in clinical environments where both types of information are crucial. The stability of ROUGE-L scores across different values of \( \alpha \), as shown in Fig~\ref{fig:alpha_performance}, supports this choice by indicating a consistent capture of relevant content regardless of slight variations in expression or phrasing. Therefore, \( \alpha = 0.5 \) represents a strategic decision to optimize the overall efficacy and reliability of the model in real-world applications. For a detailed mathematical formulation of how \( \alpha \) influences layer weighting, refer to Equation~\ref{eq:weighting_scheme} where we define the weighting scheme across the layers.

\section{Additional Experiments}
\label{appendix:additional}
To better understand how different generation settings and model improvements affect performance, we conduct additional experiments across a range of token limits and temperature settings. Table~\ref{tab:performance-metrics} presents performance metrics for all model variants. The results show that the inclusion of Context-Preserving Token Filtering (CPTF) consistently improves LLaMA 3.2 across all configurations, while the ConTextual model achieves the best performance on several metrics, particularly at higher token budgets. Notably, the highest BERT Precision (84.09) is achieved by ConTextual at Token=100, Temp=0.1, while the highest ROUGE-L (11.04) and B-2 (4.65) scores are observed at Token=300, Temp=0.1, indicating strong performance with extended generation. These findings suggest that both architectural modifications and generation hyperparameters play a critical role in optimizing clinical summarization quality.

The temperature parameter, $T$, was varied across \{0.1, 0.7, 0.9\}, influencing the stochasticity of token sampling. Lower values ($T=0.1$) enforce deterministic outputs, ensuring high precision in extracted entities, while higher values ($T=0.9$) promote diversity in text generation, potentially capturing a broader semantic spectrum. Similarly, the maximum token constraint was adjusted across \{100, 200, 300\}, enabling a direct investigation of sequence length on linguistic coherence and computational feasibility.
\begin{table*}[h!]
    \centering
    \caption{\textbf{Performance Metrics for Model Variants.} Higher values indicate better performance for all metrics (↑). The best result for each metric is highlighted in \textbf{bold}.}
    \label{tab:performance-metrics}
    \tiny
    \setlength{\tabcolsep}{5pt}
    \renewcommand{\arraystretch}{0.85}
    \begin{tabular}{@{}p{2.4cm}cc|cccccc@{}}
        \toprule
        \textbf{Model} & \textbf{Token} & \textbf{Temp} & \textbf{B-1 (↑)} & \textbf{B-2 (↑)} & \textbf{R-L (↑)} & \textbf{BERT-P (↑)} & \textbf{BERT-R (↑)} & \textbf{BERT-F1 (↑)} \\
        \midrule
        \multirow{9}{*}{\raggedright LLaMA 3.2} 
        & 100 & 0.1 & 10.92 & 5.45 & 7.37 & 79.58 & 80.23 & 79.89 \\
        &     & 0.7 & 10.90 & 5.40 & 7.33 & 79.58 & 80.23 & 79.89 \\
        &     & 0.9 & 10.91 & 5.41 & 7.32 & 79.58 & 80.23 & 79.89 \\
        & 200 & 0.1 & 10.76 & 5.37 & 7.33 & 79.58 & 80.23 & 79.89 \\
        &     & 0.7 & 10.73 & 5.35 & 7.24 & 79.58 & 80.23 & 79.89 \\
        &     & 0.9 & 10.73 & 5.34 & 7.25 & 79.58 & 80.23 & 79.89 \\
        & 300 & 0.1 & 10.70 & 5.34 & 7.28 & 79.58 & 80.23 & 79.89 \\
        &     & 0.7 & 10.68 & 5.33 & 7.26 & 79.58 & 80.23 & 79.89 \\
        &     & 0.9 & 10.64 & 5.30 & 7.20 & 79.58 & 80.23 & 79.89 \\
        \midrule
        \multirow{9}{*}{\raggedright LLaMA 3.2 + CPTF} 
        & 100 & 0.1 & 15.12 & 6.75 & 8.92 & 80.39 & 81.63 & 80.99 \\
        &     & 0.7 & 15.25 & 6.81 & 8.99 & 80.39 & 81.63 & 80.99 \\
        &     & 0.9 & \textbf{15.28} & 6.79 & 9.00 & 80.39 & 81.63 & 80.99 \\
        & 200 & 0.1 & 14.63 & 6.52 & 8.74 & 80.39 & 81.63 & 80.99 \\
        &     & 0.7 & 14.95 & 6.65 & 8.88 & 80.39 & 81.63 & 80.99 \\
        &     & 0.9 & 15.10 & 6.72 & 8.93 & 80.39 & 81.63 & 80.99 \\
        & 300 & 0.1 & 14.26 & 6.36 & 8.58 & 80.39 & 81.63 & 80.99 \\
        &     & 0.7 & 14.75 & 6.58 & 8.80 & 80.39 & \textbf{81.63} & 80.99 \\
        &     & 0.9 & 14.82 & 6.54 & 8.77 & 80.39 & 81.63 & 80.99 \\
        \midrule
        \multirow{9}{*}{\raggedright ConTextual} 
        & 100 & 0.1 & 3.80 & 1.48 & 8.73 & \textbf{84.09} & 79.30 & 81.36 \\
        &     & 0.7 & 3.60 & 1.30 & 8.30 & 83.76 & 79.29 & 81.44 \\
        &     & 0.9 & 3.24 & 1.21 & 7.83 & 83.43 & 79.11 & 81.19 \\
        & 200 & 0.1 & 9.49 & 3.60 & 10.68 & 83.05 & 80.35 & \textbf{81.65} \\
        &     & 0.7 & 9.06 & 3.35 & 9.98 & 82.72 & 80.32 & 81.48 \\
        &     & 0.9 & 8.64 & 2.90 & 9.33 & 82.23 & 80.18 & 81.18 \\
        & 300 & 0.1 & 12.63 & \textbf{4.65} & \textbf{11.04} & 82.19 & 80.61 & 81.37 \\
        &     & 0.7 & 12.17 & 4.26 & 10.37 & 82.11 & 80.66 & 81.36 \\
        &     & 0.9 & 11.96 & 3.73 & 9.83 & 81.66 & 80.48 & 81.05 \\
        \bottomrule
    \end{tabular}
\end{table*}

\section{CPTF Workflow Example}
\label{appendix:cptf}
To demonstrate the practical application and efficacy of the proposed Context-Preserving Token Filtering (CPTF) mechanism, we present an example workflow that illustrates the transformation of a verbose clinical note into a reduced, semantically significant representation and, finally, into a concise summary. This process highlights the capability of CPTF to optimize input text for efficient processing while retaining clinically relevant information.

\begin{longtable}{|p{0.2\textwidth}|p{0.7\textwidth}|}
    \caption{\textbf{Example} CPTF Workflow. From Original Clinical Note to Final Summary. This workflow demonstrates how CPTF reduces verbosity while retaining clinically relevant insights.\label{tab:example-workflow}}\\
    \hline
    \textbf{Stage} & \textbf{Content} \\ \hline
    \endfirsthead
    
    \multicolumn{2}{c}%
    {{\tablename\ \thetable{} -- continued from previous page}} \\
    \hline
    \textbf{Stage} & \textbf{Content} \\ \hline
    \endhead
    
    \hline \multicolumn{2}{|r|}{{Continued on next page}} \\ \hline
    \endfoot
    
    \hline
    \endlastfoot
    
\textbf{Input} &
\small
\textless SEX\textgreater\ M, \textless SERVICE\textgreater\ MEDICINE, \textless ALLERGIES\textgreater\ ibuprofen, \textless CHIEF COMPLAINT\textgreater\ Fever, altered mental status.  
This is a middle-aged male with a past medical history significant for ruptured AVM, status post craniotomy, and prior intracranial abscess, who is presenting today to the emergency department with fever and altered mental status. On admission, he was noted to be febrile and somewhat confused. A non-contrast CT scan of the head was performed which showed no acute intracranial abnormalities. Laboratory workup revealed elevated CRP and thrombocytosis. The patient was subsequently diagnosed with a urinary tract infection due to Klebsiella species and prostatitis. He was started on broad-spectrum antibiotics. Neurology and infectious disease teams were consulted for further management. A PET scan was obtained which was suggestive of prostatitis. \\ \hline

\textbf{CPTF Output} &
\small
\textless SERVICE\textgreater\ MEDICINE, \textless CHIEF COMPLAINT\textgreater\ Fever, altered mental status.  
Middle-aged male with history of ruptured AVM, craniotomy, and intracranial abscess presenting with fever and altered mental status. On admission, febrile and confused. CT head showed no acute intracranial abnormalities. Labs showed elevated CRP and thrombocytosis. Diagnosed with Klebsiella UTI and prostatitis. Started on antibiotics. Neurology and infectious disease consulted. PET scan suggestive of prostatitis. \\ \hline

\textbf{Summary} &
\small
Patient with ruptured AVM and intracranial abscess presented with fever and altered mental status. Imaging showed no acute intracranial changes. Labs revealed elevated CRP. Diagnosed with Klebsiella UTI and prostatitis, treated with antibiotics. Follow-up with neurology and infectious disease advised. \\ \hline

\end{longtable}

The workflow, illustrated in Table~\ref{tab:example-workflow}, demonstrates the transformation of unstructured clinical text through three distinct stages. It begins with the original clinical note, which comprises verbose and unstructured text. This input, while containing critical medical insights, is often interspersed with redundant and extraneous details that hinder computational efficiency. The Context-Preserving Token Filtering (CPTF) mechanism is then applied to process the input text, dynamically identifying and retaining semantically significant tokens essential for downstream tasks. By filtering out irrelevant and redundant information, CPTF reduces verbosity and computational overhead while preserving key clinical insights. Finally, the reduced text, enriched with contextual domain knowledge, is utilized to generate a concise and clinically actionable summary. This final output aligns with domain-specific requirements and effectively supports clinical decision-making by distilling complex narratives into precise and meaningful insights. This example highlights the role of CPTF in improving efficiency and preserving essential clinical details. By combining token filtering with the knowledge-enhanced summarization pipeline, the workflow ensures that the final output is both computationally optimized and clinically relevant.

\section{Prompting Strategies}
\label{appendix:prompting}
To guide the language model’s generation process, we construct prompts in a structured manner for each input instance. Each prompt begins with a task-specific instruction that explicitly defines the summarization objective, optionally includes exemplar demonstrations (in one-shot or few-shot settings), and concludes with the clinical input to be summarized. This design grounds the model’s output in domain-specific linguistic and clinical conventions while retaining the flexibility to accommodate the variability inherent in clinical narratives. Integrated into our framework, this prompting strategy significantly improves the model’s ability to produce coherent, concise, and clinically actionable summaries that facilitate downstream tasks such as entity extraction and structured knowledge graph construction. We evaluate three prompting paradigms—zero-shot, one-shot, and few-shot—using standard summarization metrics, as detailed in Table~\ref{tab:prompting-strategies}.
Empirically, few-shot prompting consistently outperforms both zero-shot and one-shot configurations across all evaluation metrics. In particular, higher ROUGE-L scores indicate improved lexical and structural alignment with reference summaries. Similarly, BERT-based metrics reveal superior F1 scores, reflecting a more effective balance between semantic precision and recall.

These results are consistent with prior findings in in-context learning, demonstrating that incorporating a small number of representative exemplars enhances the model’s generalization capabilities. Our findings underscore the importance of carefully engineered prompts in optimizing language model performance on clinical summarization tasks.

\begin{table*}[ht]
\caption{\textbf{Performance Comparison of Different Prompting Strategies.} Results show that few-shot prompting achieves superior performance on ROUGE-L and BERT metrics compared to zero-shot and one-shot approaches. Higher values indicate better performance across all metrics.}
\label{tab:prompting-strategies}
\centering
\footnotesize
\resizebox{\textwidth}{!}{ 
\begin{tabular}{@{}l|ccc|ccc@{}}
\toprule
\multirow{2}{*}{\textbf{Prompting Strategy}} & \multicolumn{3}{c|}{\textbf{Lexical Alignment}} & \multicolumn{3}{c}{\textbf{Semantic Alignment}} \\
\cmidrule(lr){2-4} \cmidrule(lr){5-7}
 & \textbf{BLEU-1 (\%)} & \textbf{BLEU-2 (\%)} & \textbf{ROUGE-L (\%)} & \textbf{BERT-P (\%)} & \textbf{BERT-R (\%)} & \textbf{BERT-F1 (\%)} \\
\midrule
Zero-shot & 11.85 & 6.41 & 8.59 & 79.83 & 81.95 & 80.58 \\
One-shot & 12.99 & 6.17 & 8.12 & 79.85 & 81.69 & 80.74 \\
Few-shot & \textbf{12.63} & \textbf{4.65} & \textbf{11.04} & \textbf{82.19} & 80.61 & \textbf{81.36} \\
\bottomrule
\end{tabular}
}
\end{table*}

\section{Few-Shot Prompt Design}
\label{appendix:Few-Shot}
For each input instance, the model is guided by a dynamically constructed prompt. The prompt begins with a clear instruction, contextualizing the task as a clinical summarization problem. It incorporates curated examples as demonstrations of the desired output style, concluding with the specific input instance requiring summarization. This structured approach transitions seamlessly from exemplar summaries to the new input, providing the model with implicit guidelines for the task.

\begin{longtable}
{@{}p{0.26\textwidth}p{0.65\textwidth}@{}}
\captionsetup{width=0.9\textwidth}
\caption{Few-Shot Prompt Template for Clinical Summarization. Each example demonstrates the input structure, task-specific context, and desired output style. These examples guide the model in generating high-quality clinical summaries for unseen instances.}\label{tab:prompt-example}\\
\toprule
\textbf{Component} & \textbf{Example: Oncology} \\
\endfirsthead
\toprule
\textbf{Component} & \textbf{Example: Oncology} \\
\endhead
\hline
\multicolumn{2}{r}{{Continued on next page}} \\
\endfoot
\endlastfoot
\midrule
\textbf{Instruction} & Summarize the provided clinical notes to produce a concise, domain-specific summary. Focus on clinically relevant information while omitting redundant details. \\
\textbf{Input} & 
45-year-old female with stage IV metastatic breast cancer. \newline
Chief Complaint: Severe thoracic back pain. \newline
\textbf{Imaging:} MRI spine reveals T5-T7 vertebral compression fractures; PET-CT shows multiple bone metastases. \newline
\textbf{Treatments:} Morphine IV PCA for pain management, radiation oncology consult, zoledronic acid 4mg IV for bone metastases, continued letrozole 2.5mg daily. \newline
\textbf{Labs:} CA 15-3: 68 U/mL (elevated), alkaline phosphatase: 220 U/L. \\
\textbf{Target Summary} & Patient with metastatic breast cancer underwent comprehensive pain management and palliative interventions, including morphine PCA, radiation consultation, and bone-targeted therapy. \\
\midrule
\textbf{Component} & \textbf{Example: Cardiology} \\
\midrule
\textbf{Instruction} & Summarize the provided clinical notes to generate a focused cardiology case summary. \\
\textbf{Input} & 
55-year-old male with history of hypertension and smoking. \newline
Chief Complaint: Acute chest pain radiating to left arm. \newline
\textbf{Diagnostics:} ECG shows ST-segment elevation in inferior leads; Troponin I: 12.4 ng/mL (significantly elevated); Cardiac ultrasound reveals anterior wall hypokinesis. \newline
\textbf{Interventions:} Immediate cardiac catheterization, primary PCI to right coronary artery, drug-eluting stent placement. \newline
\textbf{Medications:} Aspirin 325mg, atorvastatin 80mg, metoprolol 25mg. \newline
\textbf{Labs:} CK-MB: 22.5 ng/mL, LDL: 142 mg/dL. \\
\textbf{Target Summary} & Patient diagnosed with acute myocardial infarction underwent immediate primary percutaneous coronary intervention with right coronary artery stenting and initiated comprehensive cardiac medical management. \\
\midrule
\textbf{Component} & \textbf{Example: Internal Medicine} \\
\midrule
\textbf{Instruction} & Summarize the provided clinical notes to generate a concise and medically accurate case summary. \\
\textbf{Input} & 
65-year-old female with severe COPD and 40-pack-year smoking history. \newline
Chief Complaint: Acute respiratory distress. \newline
\textbf{Diagnostics:} Chest X-ray shows bilateral hyperinflation; ABG: pH 7.32, PaCO2 65 mmHg; Spirometry: FEV1 32\% predicted. \newline
\textbf{Interventions:} Non-invasive ventilation, IV methylprednisolone 125mg, nebulized albuterol and ipratropium. \newline
\textbf{Medications:} Prednisone 40mg daily, azithromycin 500mg. \newline
\textbf{Labs:} WBC: 14,200/µL, sputum culture: Pseudomonas aeruginosa. \\
\textbf{Target Summary} & \textit{(To be generated)} \\
\bottomrule
\end{longtable}

The design emphasizes domain-specific contextualization by leveraging structured tags that align with clinical documentation conventions. The curated few-shot examples, as illustrated in Table~\ref{tab:prompt-example}, act as semantic anchors, enabling the model to adapt effectively to the intricacies of medical narratives. This approach dynamically accommodates the variability inherent in clinical notes while preserving consistency across varying inputs. By grounding the language model’s generative capabilities in domain-specific examples, the prompt design facilitates the production of high-quality, semantically faithful summaries. When integrated into our Knowledge Graph-Enhanced Attention-Guided Summarization pipeline, this methodology enhances the extraction of actionable insights from complex medical narratives, advancing both the accuracy and utility of clinical summarization.

\section{Model Selection}
\label{appendix:Model}
LLaMA 3.2 1B was chosen due to its popularity as a lightweight yet effective foundation model. Developed by Meta AI, LLaMA (Large Language Model Meta AI) is a family of transformer-based autoregressive models optimized for efficiency and scalability. The 1B variant is designed for resource-constrained environments while maintaining strong performance across various NLP tasks. Unlike traditional large-scale models, LLaMA 3.2 1B prioritizes computational efficiency, making it a suitable choice for real-world applications where inference speed and accessibility are critical factors. The model's architecture incorporates improvements in tokenization, attention mechanisms, and optimization techniques, enabling it to handle text generation tasks effectively with a relatively small parameter count. Its strong performance in retrieval-augmented and few-shot learning scenarios further supports its applicability in domains requiring efficient text processing. Given these characteristics, LLaMA 3.2 1B serves as a representative baseline for evaluating the effectiveness of lightweight language models in summarization tasks.

\section{Use Case: Clinical Trial Recruitment}
\label{appendix:Use Case}
Efficient participant identification remains a major bottleneck in clinical trial recruitment. Key eligibility criteria—such as age, comorbidities, prior treatments, and disease progression—are typically embedded in unstructured clinical narratives, including discharge summaries and progress notes. The heterogeneous and verbose nature of these records makes manual review time-consuming and error-prone. We address this challenge by introducing a framework that leverages summarization as a preprocessing step to distill unstructured texts into concise, semantically rich representations. Central to this approach is \textit{Context-Preserving Token Filtering (CPTF)}, which selectively retains clinically salient information. The resulting summaries are then aligned with a domain-specific \textit{Knowledge Graph (KG)} that encodes structured relationships among clinical entities (e.g., diagnoses, medications, outcomes).
In the context of a multiple sclerosis (MS) clinical trial, our framework automates the identification of relevant patient characteristics—such as relapse history or immunotherapy exposure—by first summarizing clinical notes and then validating these attributes through KG-based reasoning. This process minimizes manual effort, improves consistency, and accelerates recruitment.
Summarization thus serves as a critical abstraction layer, transforming unstructured narratives into actionable representations. Combined with KG integration, the framework enables scalable and accurate patient screening, even in resource-limited settings, and illustrates the broader applicability of summarization-driven workflows in clinical research.

\end{document}